\documentclass[journal]{IEEEtran}

\usepackage{xcolor,soul,framed} 

\colorlet{shadecolor}{yellow}
\usepackage[pdftex]{graphicx}
\graphicspath{{../pdf/}{../jpeg/}}
\DeclareGraphicsExtensions{.pdf,.jpeg,.png}

\usepackage[cmex10]{amsmath}
\usepackage{array}
\usepackage{mdwmath}
\usepackage{mdwtab}
\usepackage{eqparbox}
\usepackage{amsfonts,amssymb}
\usepackage{url}
\usepackage{cite} 
\usepackage{multirow}
\usepackage{multicol}

\begin{document}
\bstctlcite{IEEEexample:BSTcontrol}
    \title{Few-Shot Rotation-Invariant Aerial Image Semantic Segmentation }
  \author{Qinglong~Cao,~\IEEEmembership{Student Member,~IEEE,}
      Yuntian~Chen,~\IEEEmembership{Member,~IEEE,}\\
      Chao~Ma,~\IEEEmembership{Member,~IEEE,}
      and~Xiaokang~Yang,~\IEEEmembership{Fellow,~IEEE}

  \thanks{This work was supported in part by the National Science Foundation of China under Grants 62106116.}
  \thanks{Q. Cao is with the MoE Key Lab of Artificial Intelligence, AI Institute, Shanghai Jiao Tong University, Shanghai 200240, China, and Eastern Institute for Advanced Study, Zhejiang, China (e-mail: caoql2022@sjtu.edu.cn).}
  \thanks{Y. Chen is with Eastern Institute for Advanced Study, Zhejiang, China, and School of Electronic Information and Electrical Engineering, Shanghai Jiao Tong University 200240, China (e-mail: ychen@eias.ac.cn).}%
  \thanks{C. Ma and X. Yang are with the MoE Key Lab of Artificial Intelligence, AI Institute, Shanghai Jiao Tong University, Shanghai 200240, China.}
  
  \thanks{\textit{Corresponding author: Yuntian Chen}}
  }

\markboth{IEEE TRANSACTIONS ON GEOSCIENCE AND REMOTE SENSING
}{Cao \MakeLowercase{\textit{et al.}}:Few-shot Rotation-Invariant Aerial Image Semantic Segmentation}

\maketitle

\begin{abstract}
Few-shot aerial image semantic segmentation is a challenging task that requires precisely parsing unseen-category objects in query aerial images with limited annotated support aerial images. Formally, category prototypes would be extracted from support samples to segment query images in a pixel-to--pixel matching manner. However, aerial objects in aerial images are often distributed with arbitrary orientations, and varying orientations could cause a dramatic feature change.  This unique property of aerial images renders conventional matching manner without consideration of orientations fails to activate same-category objects with different orientations. Furthermore, the oscillation of the confidence scores in existing rotation-insensitive algorithms, engendered by the striking changes of object orientations, often leads to false recognition of lower-scored rotated semantic objects. To tackle these challenges, inspired by the intrinsic rotation invariance in aerial images, we propose a novel few-shot rotation-invariant aerial semantic segmentation network (FRINet) to efficiently segment aerial semantic objects with diverse orientations. Specifically, through extracting orientation-varying yet category-consistent support information, FRINet provides  rotation-adaptive matching for each query feature in a feature-aggregation manner. Meanwhile, to encourage consistent predictions for aerial objects with arbitrary orientations, segmentation predictions from different orientations are supervised by the same label and further fused to obtain the final rotation-invariant prediction in a complementary manner. Moreover, aiming at providing a better solution space, the backbones are newly pre-trained in the base category to basically  boost the segmentation performance.  Extensive experiments on the few-shot aerial semantic segmentation benchmark  demonstrate  that  the proposed FRINet achieves a  new state-of-the-art performance. The code is available at https://github.com/caoql98/FRINet.

\end{abstract}


%
\IEEEpeerreviewmaketitle

\section{Introduction}
\IEEEPARstart{S}{emantic}  segmentation of remote sensing images is an indispensable task in earth vision with a broad range of applications, such as surveillance of urban areas~\cite{forster1985examination,jat2008monitoring,ru2021land}, building/road detection~\cite{ayala2021deep,qiao2023weakly,ding2021non}, and traffic management~\cite{zhao2022satsot,macioszek2021extracting,jian2019combining}. With the rapid development of deep learning technology,  fully supervised segmentation algorithms have shown impressive performance on aerial image semantic segmentation~\cite{chai2020aerial,deng2021ccanet,luo2019high,zheng2021entropy,zhou2023weakly}. Yet the success of deep learning-based  methods heavily depends on the availability of large-scale annotated datasets~\cite{chaudhuri2017multilabel,shao2020multilabel,waqas2019isaid}  that are time-consuming and laborious to construct, and it is often infeasible to obtain annotated data for all semantic categories. Additionally, current fully supervised algorithms fail to handle the novel categories with limited annotated samples, resulting in poor performance~\cite{yao2021scale}. To address these challenges, it is crucial to study the few-shot aerial image semantic segmentation task, which requires accurately parsing query aerial images with only a few annotated support samples.
\begin{figure}
  \begin{center}
  \includegraphics[width=1.0\linewidth]{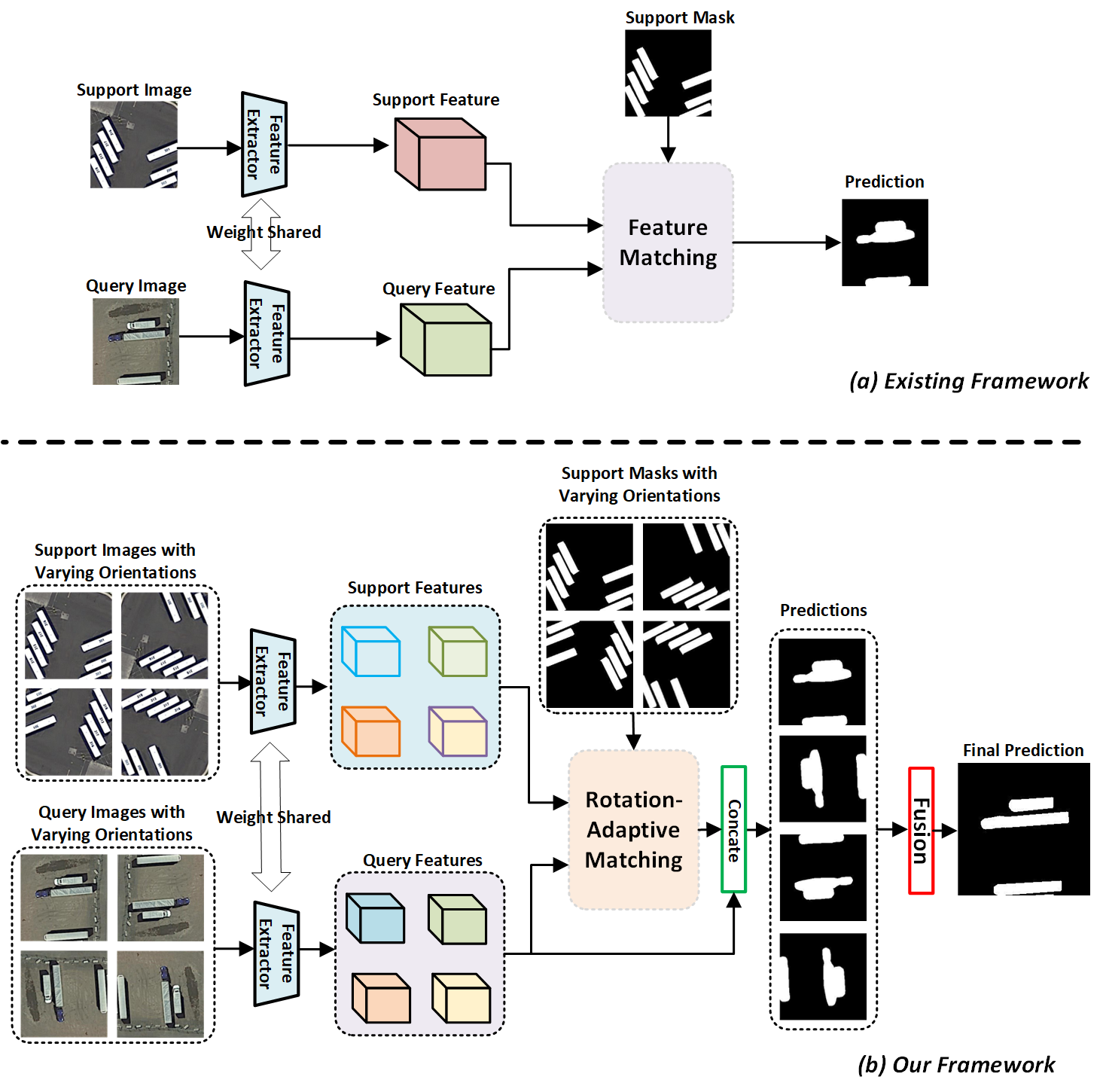}\\
  \caption{The comparison between the existing FSS framework for aerial images  and our
 proposed few-shot rotation invariant semantic segmentation framework.}\label{fig1}
  \end{center}
\end{figure}

Few-shot semantic segmentation (FSS) algorithms~\cite{zhang2020sg,wang2019panet,zhang2019canet,dong2018few,xie2021few,tian2020prior,lang2023hpa,lang2023bam} typically begin by generating category prototypes from the extracted support features. These prototypes are then utilized to accurately segment query images in a pixel-to-pixel matching manner. Following this pipeline, many advanced FSS algorithms have been proposed to boost segmentation performance via stronger category prototypes~\cite{wang2019panet,fan2022self}, better matching strategies~\cite{yang2020prototype,wang2020few}, or distracting information elimination~\cite{liu2022learning,liu2022intermediate}.  Recently, aiming at dealing with the large variance of aerial objects’ appearances and scales, Yao \textit{et al.}~\cite{yao2021scale} proposed the first few-shot aerial image semantic segmentation algorithm to provide a scale-aware detailed matching.

Previous simple matching frameworks directly expanded category prototypes as  matching features and further concatenated these matching features with query features to activate category-related semantic objects. However, many semantic objects with the same category in aerial images usually appear with arbitrary orientations, resulting in a dramatic class-agnostic feature change. The existing simple matching manner clearly could not tackle such a dramatic class-agnostic feature change, and same-category objects with different orientations might not be activated. Similarly, due to the consistent negligence of the critical rotation invariance in aerial images, current rotation-insensitive FSS solutions~(Figure \ref{fig1} (a)) have been prone to  unstable aerial image segmentation, where  aerial semantic objects' confidence scores can see drastic oscillation with the change of orientation, and the rotated semantic objects with low confidence scores may be regarded as background.

To tackle these problems, in this paper, we propose a fresh few-shot rotation-invariant aerial semantic segmentation network (FRINet) to efficiently segment aerial semantic objects with diverse orientations.  The design of the proposed FRINet (Figure \ref{fig1} (b)) is inspired by human knowledge, i.e., the category of objects in aerial images remains consistent after arbitrary rotation. This knowledge could work as an implicit constraint for rotation-invariance learning. Particularly, for the support images, the aerial objects distributed with different orientations could provide orientation-varying but category-consistent support information that enables the network to achieve orientation-adaptive matching. For the query images, the segmentation model should make consistent predictions for the aerial objects distributed with varying orientations, which could encourage the network to make a steady aerial image semantic segmentation.

More specifically, FRINet first extracts support features and query features, respectively for support and query images with varying orientations.  Then, through mask averaging pooling, the corresponding support masks are applied to the support features to generate category-consistent support prototypes with diverse orientations. To provide orientation-adaptive matching for query features, the relation scores between each element of query features and these orientation-varying support prototypes are computed. Utilizing these relation scores as the aggregating scores, each element of query features could obtain an orientation-adaptive prototype. By correspondingly concatenating these prototypes with the query features, aerial semantic objects with different orientations could be adaptively activated. Subsequently,  aiming at encouraging the segmentation model to make consistent predictions from different orientations, the predictions from different orientation branches are supervised with the same rotated labels. In this way, the same-category objects with different rotations would be pulled closer in the embedding space, and the foreground could be searched in varying rotations. Finally,  by aggregating these predictions from different  orientations to generate the final prediction in a complementary manner, the iterative meta-training process could provide abundant complementary visual patterns to develop a powerful few-shot rotation-invariant segmentation network. 

Moreover, existing FSS solutions tend to leverage backbones (e.g. VGG16~\cite{simonyan2014very}, Resnet50~\cite{he2016deep}) pre-trained in the large-scale ImageNet~\cite{deng2009imagenet} as the feature extractor for both support and query images. Yet these pre-trained backbones working for classification could only provide a global perceptive feature space for support and query images, which clearly cannot satisfy the pixel-level judge requirement for the segmentation task. To provide more rational and detailed embedding space, the backbones shall first be pre-trained in the base category as the feature extractor for the segmentation task. 

By elaborately leveraging category-invariance in rotation,  the proposed FRINet provides a rotation-adaptive matching and further performs a steady segmentation process with the rotation-consistent constraint. FRINet successfully addresses previous challenges via rotation-invariant  learning  and achieves state-of-the-art segmentation performance. The main contributions of this paper are summarized as follows:

\begin{itemize}
	\item To the best of our knowledge, we present the first attempt to build a rotation-invariant aerial semantic segmentation network under the few-shot setting, and the proposed network is able to provide a rotation-consistent prediction in a rotation-adaptive matching manner.  
	
	\item Aiming at constructing a more rational and detailed feature space for FSS in aerial images, we newly provide backbones pre-trained on the base category to boost the segmentation performance to a higher level.
	
	\item Experimental results on the FSS benchmark for aerial images demonstrate that the proposed FRINet outperforms existing state-of-the-art  few-shot aerial segmentation methods by a large margin.
	
\end{itemize}

\section{Related Work}
Semantic segmentation has been extensively studied for the last
few decades. In this section, we will first review the related works with regard to aerial image semantic segmentation and then introduce advanced few-shot semantic segmentation algorithms. 
\subsection {Aerial Image Semantic Segmentation}
Aerial image semantic segmentation aims to precisely classify each pixel of aerial images. Recently, with the development of deep learning technology, many advanced aerial image semantic segmentation algorithms have been proposed~\cite{luo2019high,kampffmeyer2016semantic,diakogiannis2020resunet,mou2019relation,ding2020lanet,li2020scattnet,niu2021hybrid}. For instance, Kampffmeyer~\textit{et al.}~\cite{kampffmeyer2016semantic} first introduces the deep convolutional neural networks (CNNs) to parse the aerial images in an uncertainty quantifying manner. By introducing residual connections~\cite{he2016deep}, atrous convolutions~\cite{chen2017deeplab}, pyramid scene parsing pooling~\cite{zhao2017pyramid}, and multi-tasking inference into previous CNNs, Diakogiannis~\textit{et al.}~\cite{diakogiannis2020resunet} propose a reliable segmentation framework for very high resolution aerial images. Following this pipeline, Mou~\textit{et al.}~\cite{mou2019relation} propose the global relation modeling to mine the  long-range spatial relations in aerial images. By introducing the popular attention mechanism into the aerial image semantic segmentation task,  Ding~\textit{et al.}~\cite{ding2020lanet} construct local attention embedding to boost the segmentation performance. Similarly, SCAttNet~\cite{li2020scattnet} leverages the popular spatial and channel attention mechanisms to handle the segmentation of aerial images. Based on previous attention mechanisms, HMANet~\cite{niu2021hybrid} proposes  hybrid multiple attention to adaptively capture global correlations in a more effective and efficient manner. Aiming at dealing with aerial objects of various sizes and materials, AFNet~\cite{liu2020afnet} builds a novel adaptive fusion network where features from different layers are adaptively selected to handle objects with different sizes. Moreover, some attempt to introduce extra information to boost segmentation performance. For example, utilizing the boundary as a supervision label,  Bokhovkin~\textit{et al.}~\cite{bokhovkin2019boundary} introduce boundary loss to enhance the feature extraction ability. Simultaneously, Chai~\textit{et al.}~\cite{chai2020aerial} compute category distance maps as segmentation guidance to respectively perform segmentation for each category. Viewing the semantic propagation as the point-wise flow, Li~\textit{et al.}~\cite{li2021pointflow} compute a sparse affinity map through adjacent features to provide a stronger feature propagation process. Recently, the famous transformer has also been introduced into aerial image semantic segmentation and has achieved promising success ~\cite{gao2021stransfuse, wang2022novel,yan2022efficient,he2022swin,wang2022unetformer,zhao2021memory}. Among these transformer-based works, to acquire better semantic embeddings, Gao~\textit{et al.}~\cite{gao2021stransfuse} jointly adopt the transformer and CNNs to extract coarse-grained and fine-grained feature representations at various semantic scales.
\subsection {Few-shot Semantic Segmentation}
The goal of few-shot semantic segmentation is to parse the unseen-category objects with only a few annotated support samples. The first attempt for FSS~\cite{dong2018few} is to leverage the category prototypes learned from the support samples to parse the category-related objects in a metric-learning manner. To provide strong category prototypes with stronger  discriminative ability, prototype alignment regularization between support and query is introduced into PANet~\cite{wang2019panet}. Simultaneously, aiming at refining the initial coarse predictions, CANet~\cite{zhang2019canet} utilizes multi-level feature comparison to iteratively refine the segmentation results. Previous works only provided a single prototype for each category. To provide more detailed category information, PMMs~\cite{yang2020prototype} introduce prototype mixture models to enforce the prototype-based semantic representation. Similarly,  Liu~\textit{et al.}~\cite{liu2020part} adopt the super-pixels concept to generate part-aware prototypes and further design a graph neural network model to perform segmentation of query images. Though previous FSS methods have acquired good segmentation performance, these algorithms ignore the critical generalization problem between novel and base categories. Focusing on tackling this issue, PFEnet~\cite{tian2020prior} leverages high-level semantic information as prior knowledge to tackle the spatial inconsistency between query and support targets.  Accurately analyzing the relations between query and support features is the key factor of FSS. Inspired by this, HSNet~\cite{min2021hypercorrelation} designs efficient 4D convolutions to mine the e fine-grained correspondence relations between the query and the support images. Furthermore, efficiently removing distracting objects is also an appropriate method to boost the segmentation performance. BAM~\cite{lang2022learning} and NERTNet~\cite{liu2022learning} adopt this concept, and respectively, eliminate the distracting objects with a base learner and a background mining loss. However, previous methods all focused on natural images and thus could not handle aerial images. Recently, aiming at providing detailed support guidance for the aerial images, SDM~\cite{yao2021scale} was first proposed to tackle few-shot aerial image semantic segmentation in a detailed matching manner. However, previous FSS solutions all ignore that the aerial objects in aerial images are often distributed with arbitrary orientations and varying orientations could cause a dramatic feature change. This property would lead to false recognition of orientation-varying aerial objects. Though the random rotation strategy could provide aerial objects with varying orientations in long-term training. only a single-orientation view is leveraged in the query-support matching for an iteration. The orientation-varying objects could still not be activated and
be falsely recognized. To conquer this challenge,  a novel
few-shot rotation-invariant aerial semantic segmentation network is first proposed in this paper.

\section{Proposed Method}

\subsection{Problem Definition}
Few-shot aerial image semantic segmentation aims to parse novel-category objects in query aerial images with only a few annotated support aerial samples. Normally, episode-based meta-learning is adopted to perform the model training. The categories of datasets would be firstly divided into two non-overlapping subsets, namely $C_{base}$ and $C_{novel}$. Subsequently, the aerial images containing base-category objects would be collected to construct the training dataset $D_{train}$, and the aerial images containing novel-category objects would be accumulated to build the test dataset $D_{test}$. Formally, given $K$-shot setting, during the training phase, $K+1$ labeled aerial images $\left\{ (I_s^1,M_s^1),(I_s^2,M_s^2), \cdot  \cdot  \cdot (I_s^k,M_s^k),({I_q},{M_q})\right\}$ of targeted category would be sampled from the  $D_{train}$ episodically, where $(I_s^i,M_s^i)$  denotes the support image-mask pair and $({I_q},{M_q})$ denotes the query image-mask pair. The goal of the few-shot aerial image segmentation model is to learn how to precisely parse the query image under the guidance of $K$-shot annotated support samples. The predicted mask $\widetilde M_q$ is supervised by the ground truth mask $M_q$. Similarly, during the test stage,  for each novel category, $K$ shot annotated support samples would be sampled from the $C_{novel}$ to inference the semantic objects of the query images. 

\begin{figure*}[t]
  \begin{center}
  \includegraphics[width=1.0\linewidth]{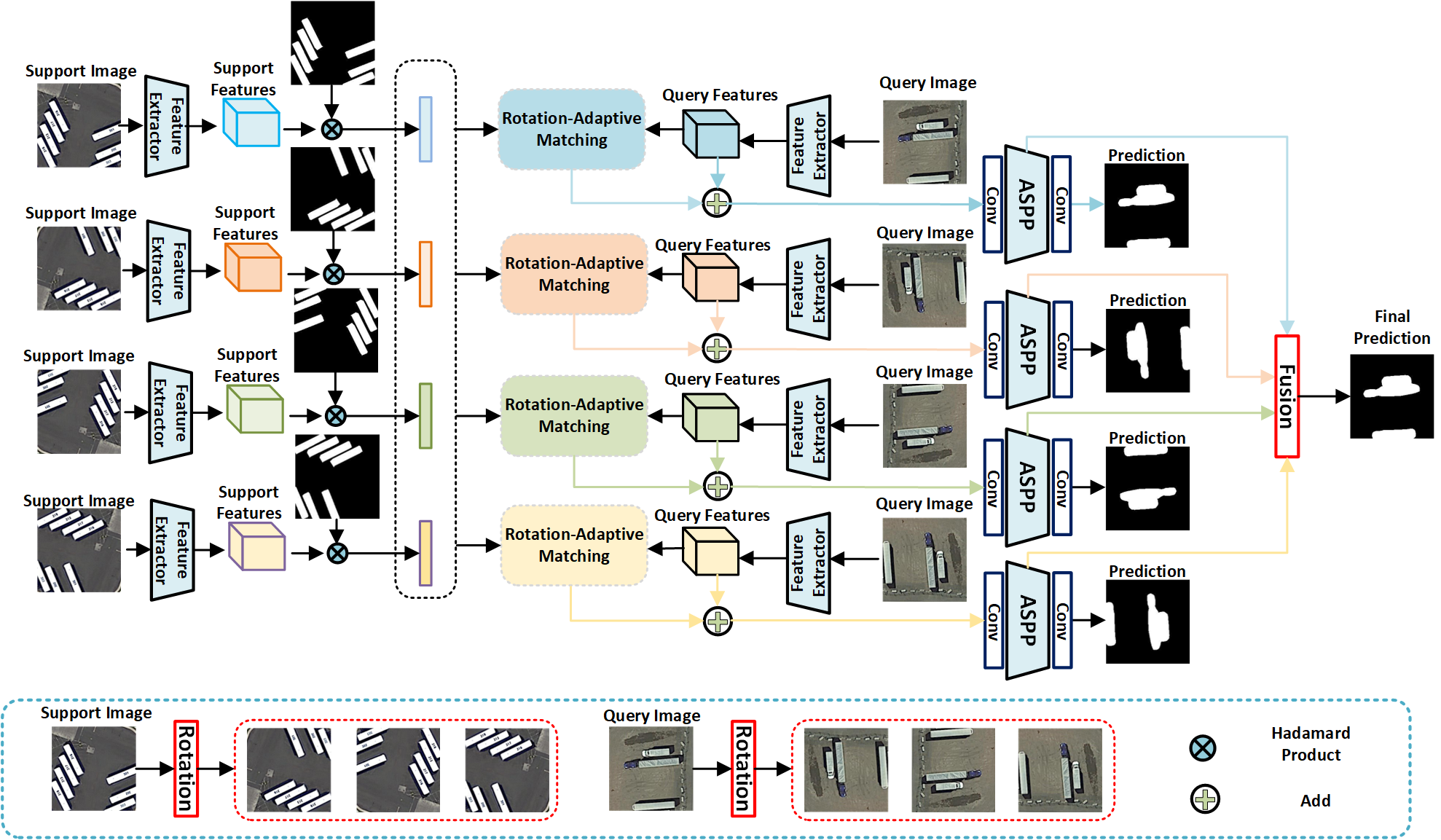}\\
  \caption{The overall framework of the proposed FRINet. Original support and query images are first rotated to obtain support and query images with varying orientations. Then the segmentation-pretrained backbones are leveraged as the feature extractor to obtain the corresponding support and query features. Then, with the category-consistent but orientation-varying prototypes, the network provides a rotation-adaptive matching for the orientation-varying query features. To encourage the network to make orientation-consistent predictions, the segmentation results from diverse orientations are supervised by the same ground truth. Finally, the predictions from different orientations are complementarily fused to obtain the final rotation-invariant prediction. }\label{fig2}
  \end{center}
\end{figure*}

\subsection{Method Overview}
As shown in Figure \ref{fig2}, inspired by the intrinsic rotation invariance in aerial images, we propose a novel
few-shot rotation-invariant aerial semantic segmentation network (FRINet) to efficiently segment aerial semantic objects with diverse orientations. Firstly, different from previous FSS solutions that utilized backbones (VGG16 or Resnet50) pre-trained in the ImageNet as feature extractors, the backbones are pre-trained in the base category  for segmentation. Then, the pre-trained backbones (VGG16 or Resnet50) are leveraged as the feature extractor to obtain support and query features, respectively, for support and query images with varying orientations. In this manner, the newly pre-trained backbones could provide a better solution for FSS segmentation. Subsequently, with given support masks and extracted support features, the rotation-varying but category-consistent support prototypes are collected through the mask average pooling operation. To provide a rotation-adaptive matching for aerial objects with varying orientations, the relation scores between orientation-varying query features and the generated support prototypes are computed. Leveraging the relation scores as guidance, each element of query features could obtain the newly generated orientation-adaptive prototype. By correspondingly utilizing these prototypes to activate the query features, the aerial objects with varying prototypes would be adaptively activated.  We then propagate the activated features into the designed convolutional layers with the Atrous Spatial Pyramid Pooling (ASPP ~\cite{chen2017deeplab}) to predict the final parsing result. To encourage the segmentation model to make consistent predictions on orientation-varying objects, the segmentation results for query images with varying orientations are supervised by the same ground truth mask. Finally, the predictions from different orientations are complementarily fused to generate the final rotation-invariant prediction.

\subsection{Feature Extractor}
Since the core idea of FSS setting is learning how to transfer meta-knowledge from the base category to the novel category,  weights frozen backbones are normally leveraged as the feature extractor to improve the generalization ability of the segmentation model. Previous FSS methods tend to use the backbones pre-trained in the large-scale ImageNet, which work as the feature extractor for classification. However, the supervision from classification could only help the feature extractor learn a global visual pattern, this pattern clearly has limited ability to support the segmentation model in performing pixel-level classification. Thus, to provide a more detailed visual pattern for the feature extractor,  a pre-trained feature extractor for segmentation is a better solution. Thus, with the abundant annotated data for the base category, we pre-train these backbones as the feature extractor for segmentation. In this manner, the newly provided backbones could provide a better feature space for the few-shot aerial image semantic segmentation and basically boost the segmentation performance.

\subsection{Rotation-Adaptive Matching}
\begin{figure}
  \begin{center}
  \includegraphics[width=1.0\linewidth]{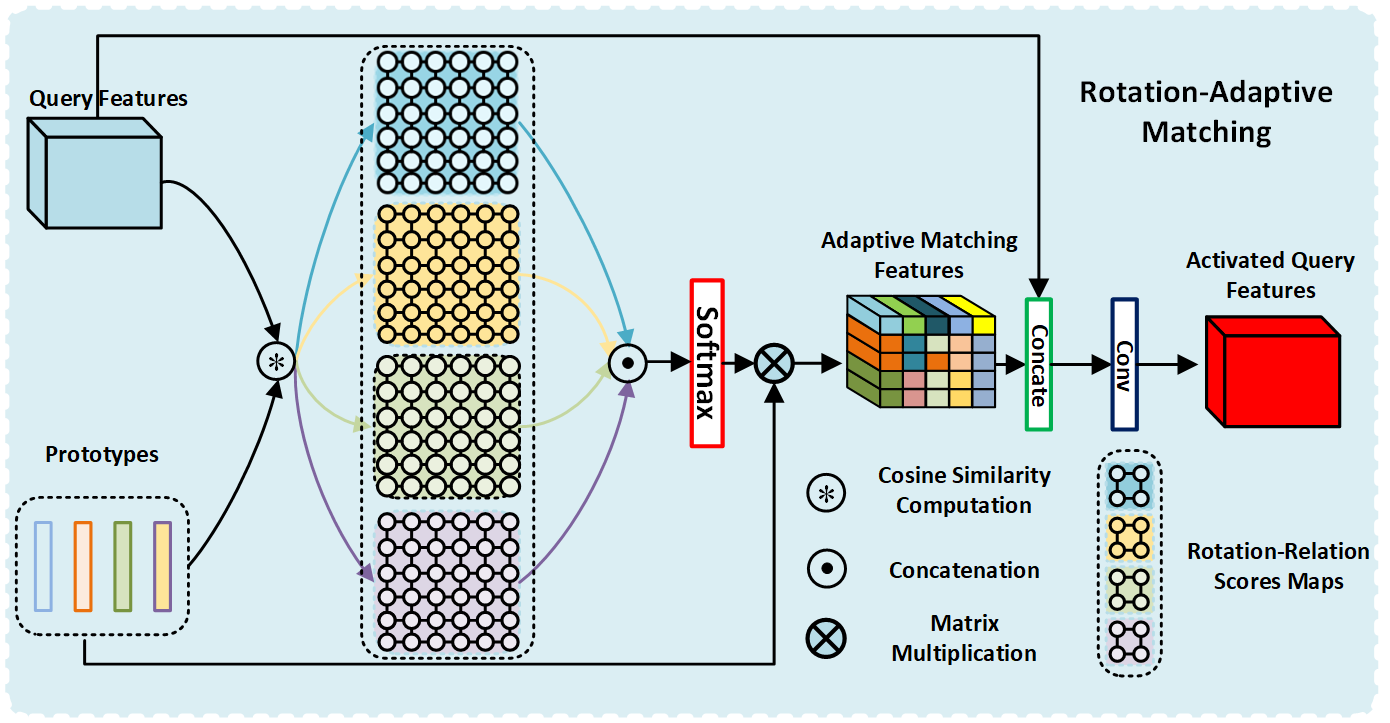}\\
  \caption{Illustration of the rotation-adaptive matching. Utilizing relation scores as guidance, the prototypes are accumulated to construct the rotation-adaptive matching features. Through concatenation and convolutional layers, rotation-adaptive matching produces the activate query features.   }\label{fig3}
  \end{center}
\end{figure}

Existing FSS methods typically focus on how to mine better category prototypes from support samples to match the query images in a pixel-to-pixel manner. However, the category-consistent semantic objects in aerial images always appear in arbitrary orientations. The previous matching paradigm that does not take orientation into account would fail to activate all aerial objects. To conquer this challenge, as shown in Figure~\ref{fig3}, we propose rotation-adaptive matching to activate the orientation-varying aerial objects. 

Assume we have the support image $I_S$ and the query image $I_Q$, the support image and query image would be firstly rotated in different angles ($90^{\circ}$, $180^{\circ}$, and $270^{\circ}$) to acquire rotated support images $\left\{I_{S1}, I_{S2}, I_{S3}\right\}$ and rotated query images $\left\{I_{Q1}, I_{Q2}, I_{Q3}\right\}$. Then, by propagating support images and query images with varying orientations, i.e.,  $\left\{I_S, I_{S1}, I_{S2}, I_{S3}\right\}$ and $\left\{I_Q, I_{Q1}, I_{Q2}, I_{Q3}\right\}$, into the pre-trained feature extractor to obtain the corresponding support features $\left\{F_s, F_{s1}, F_{s2}, I_{s3}\right\}$ and query features $\left\{F_q, F_{q1}, F_{q2}, F_{q3} \right\}$, where the size of these features equals $C \times H \times W$.

Given the support mask $M_s$, we first rotate the original mask with the same angles to obtain the corresponding masks $\left\{M_s, M_{s1}, M_{s2}, M_{s3}\right\}$. Then,  we respectively conduct masked average pooling operations on the orientation-varying support features to obtain the orientation-varying prototypes$\left\{P_s, P_{s1}, P_{s2}, P_{s3}\right\}$:
\begin{equation}\label{equation1}
\mathop {{P_i}}\limits_{i \in \{ s,s1,s2,s3\} }  = Avgpool({F_i} \odot {M_i}) \in {\mathbb{R}^{C \times 1 \times 1}}
\end{equation}

With these category-consistent prototypes with varying orientations, we could leverage the similarities between prototypes and each pixel of the query to provide the rotation-adaptive matching features in an aggregating manner. The rotation-adaptive matching is applied to all query features $\left\{F_q, F_{q1}, F_{q2}, F_{q3} \right\}$. For better understanding, we then take $F_q$ for a more detailed explanation. 

Query feature $F_q$ is first flattened in spatial dimension as query feature sets $\left\{ { l_1^q,l_2^q, \cdot  \cdot  \cdot , l_N^q} \right\}$, where $ l \in {\mathbb{R}^{C \times 1 \times 1}}$ and $N = H \times W$. Subsequently, we choose the cosine similarity as the correlation function  to compute the pixel-level relation scores between query features and orientation-varying prototypes:
\begin{equation}
\cos ({P_i},l_j^q) = \frac{{{P_i}l_j^q}}{{\left\| {{P_i}} \right\|\left\| {l_j^q} \right\|}},i \in \{ s,s1,s2,s3\} ,j \in \{ 1,2,...,N\} 
\end{equation}

By collecting relation cores for each element of query features, we could obtain the rotation-relation score maps $\left\{m_s,m_{s1},m_{s2},m_{s3} \right\} \in {\mathbb{R}^{1 \times H \times W}}$. Then by concatenating  these score maps in channel dimension, and applying softmax operation in these maps, we could acquire the rotation-relation matrix $m_p \in  \mathbb{R}^{4 \times H \times W}$ between query features and orientation-varying prototypes:
\begin{equation}
{m_p} = Softmax (Cat({m_s},{m_{s1}},{m_{s2}},{m_{s3}}))
\end{equation}

It is notable that each score in the relation matrix could denote the orientation distance of each element of query features. Thus, according to the distance, the rotation-adaptive prototypes $P_r$ could be computed for each element, by collecting these prototypes, we could obtain the rotation-adaptive matching features $F_r \in \mathbb{R}^{C \times H \times W} $. This process could be achieved by the matrix multiplication between the  orientation-varying prototypes and  rotation-relation matrix $m_p$:
\begin{equation}
{F_r} = Cat({P_s},{P_{s1}},{P_{s2}},{P_{s3}}) \otimes {M_p} 
\end{equation}
where $\otimes$ denotes  matrix multiplication.

Subsequently, to corresponding activate the aerial objects, the  rotation-adaptive matching features are concatenated with the query feature and further propagated into the convolutional layers to obtain the activated query features $F_a$:
\begin{equation}
{F_a} = Conv(Cat({F_r},{F_q}))
\end{equation}

\begin{table*}[t!]
\caption{Categories in the iSAID-$5^i$ dataset}
\resizebox{\textwidth}{!}{
\begin{tabular}{c|c|c|c|c|c|c|c|c|c|c|c|c|c|c}
\hline
C1   & C2       & C3    & C4    & C5   & C6     & C7   & C8  & C9   & C10   & C11    & C12    & C13   & C14   & C15    \\ \hline
ship & \begin{tabular}[c]{@{}c@{}}storage   \\ tank\end{tabular} & \begin{tabular}[c]{@{}c@{}}baseball   \\ diamond\end{tabular} & \begin{tabular}[c]{@{}c@{}}tennis   \\ court\end{tabular} & \begin{tabular}[c]{@{}c@{}}basketball\\    court\end{tabular} & \begin{tabular}[c]{@{}c@{}}ground   \\ track \\ field\end{tabular} & bridge & \begin{tabular}[c]{@{}c@{}}large   \\ vehicle\end{tabular} & \begin{tabular}[c]{@{}c@{}}small \\ vehicle\end{tabular} & helicopter & \begin{tabular}[c]{@{}c@{}}swimming   \\ pool\end{tabular} & roundabout & \begin{tabular}[c]{@{}c@{}}soccer   \\ ball \\ field\end{tabular} & plane & harbor \\ \hline
\end{tabular}
}
\label{class}
\end{table*}

\begin{table}[t!]
\caption{Testing classes for the three-fold  cross-validation. The training classes of  iSAID-$5^i, i=0,1,2$ are disjoint with the testing classes.}
\resizebox{\linewidth}{!}{
\begin{tabular}{c|c}
\hline
Dataset  & Test classes   \\ \hline
iSAID-$5^0$ & \begin{tabular}[c]{@{}c@{}}ship, storage tank, baseball diamond,\\  tennis court, basketball court\end{tabular} \\ \hline
iSAID-$5^1$ & \begin{tabular}[c]{@{}c@{}}ground track field, bridge, large vehicle, \\ small vehicle, helicopter\end{tabular} \\ \hline
iSAID-$5^2$ & \begin{tabular}[c]{@{}c@{}}swimming pool, roundabout, \\ soccerball field, plane, harbor\end{tabular}           \\ \hline
\end{tabular}
}
\label{splits}
\end{table}

\subsection{Rotation-Invariant Segmentation}
Aiming at encouraging the segmentation model to make consistent predictions, the segmentation results from orientation-varying query images are supervised with the same ground truth labels. Then the segmentation predictions from different orientations are complementarily fused to obtain the final rotation-invariant result. Specifically, after applying rotation-adaptive matching to query features $\left\{F_q, F_{q1}, F_{q2}, F_{q3}\right\}$, we could obtain the correspondingly activated query features $\left\{F_a, F_{a1}, F_{a2}, F_{a3}\right\}$. By propagating these activated query features into the ASPP modules and designing convolutional layers,  we could infer the segmentation results  $\left\{R_a, R_{a1}, R_{a2}, R_{a3}\right\}$:
\begin{equation}
\mathop {{R_i}}\limits_{i \in \{ a,a1,a2,a3\} }  = C{\text{onv}}(ASPP(Conv({F_i})))
\end{equation}
\begin{equation}
 \mathop {{R_i}}\limits_{i \in \left\{ {a,a1,a2,a3} \right\}}  = \left\{ {R_i^f,R_i^b} \right\}   
\end{equation}
where the ${R_i} \in {\mathbb{R}^{2 \times H \times W}}$, $R_i^f$ denotes the foreground probability map and $R_i^b$ denotes the background probability map. In order to leverage the same ground truth $M_{GT}$ as the supervision, all these segmentation results would be rotated inversely in corresponding angles:
\begin{equation}
loss_{rotation} = \sum\limits_{i \in \{ a,a1,a2,a3\} } {cross\_entropy({M_{GT}},Ro({R_i}))} 
\end{equation}
where $Ro$ denotes that rotating the segmentation results in corresponding angles. Subsequently, these segmentation results are complementarily fused to obtain the final rotation-invariant segmentation results. Particularly, through convolutional neural layers,  the foreground probability maps, and background probability maps are respectively fused to obtain the final foreground probability map and background probability map, which are further concatenated to obtain the final  rotation-invariant segmentation result $R_{F}$:
\begin{equation}
  R_{F}^f = Conv(Cat(R_a^f,Ro(R_{a1}^f),Ro(R_{a1}^f),Ro(R_{a1}^f))) 
\end{equation}
\begin{equation}
   R_{F}^f = Conv(Cat(R_a^f,Ro(R_{a1}^f),Ro(R_{a1}^f),Ro(R_{a1}^f)))   
\end{equation}
\begin{equation}
R_{F} = \left\{R_{F}^f, R_{F}^b\right\}  
\end{equation}
The final rotation-invariant segmentation result is also supervised by the ground truth mask $M_{GT}$:
\begin{equation}
los{s_{main}} = cross\_entropy({M_{GT}},{R_F})
\end{equation}
The overall supervision loss is computed as:
\begin{equation}
loss_{all} = loss_{main} + \mu  * loss_{rotation}
\end{equation}
where $\mu$ is the adjusting factor, and the factor is empirically
set as 0.25 in this paper.

\section{Experiments}
To demonstrate the effectiveness of the proposed FRINet, extensive experiments are performed on the few-shot aerial segmentation benchmark, and the introduction of this section is illustrated as follows.  We first describe the few-shot aerial segmentation benchmark and the evaluation metric. Then, the implementation details of FRINet are provided for better realization. Subsequently, the comparisons between the proposed FRINet and other advanced FSS solutions are present, and we analyze the segmentation results comprehensively.  Finally,  a series of ablation studies are conducted to illustrate the differences between FRINet and random flipping strategy, and demonstrate the impact of each component in our proposed FRINet.
\subsection{Dataset and Evaluation Metric}
We follow the experimental setting in SDM~\cite{yao2021scale} to utilize the iSAID-$5^i$ dataset as the benchmark. The iSAID-$5^i$ dataset is constructed based on the iSAID dataset~\cite{waqas2019isaid}. It mainly consists of 655451 object instances across 15 categories and the details of the categories are illustrated in Table ~\ref{class}. Particularly, for the 15 object categories in the iSAID-$5^i$ dataset, the cross-validation method is leveraged to evaluate the proposed model by sampling five classes as test categories $D_{test}$ and leveraging the left ten classes as the categories of the training set$D_{train}$. The details of the class splits are shown in Table \ref{splits}, where i is the fold number. The  iSAID-$5^i$ dataset contains 18076 images for training and 6363 images for validation, where the size of images equals $3 \times 256 \times 256$.  Following the evaluation setting in \cite{yao2021scale}, we adopt the class mean intersection over union (mIOU) as our evaluation metric, which could straightly reflect the model performance. Formally, the mIOU could be defined as follow:
\begin{equation}
mIOU = \frac{1}{C}\sum\limits_{i = 1}^C {IO{U_i}}
\end{equation}
where $C$ is the number of categories in each split and ${IO{U_i}}$ is the  intersection over union of class $i$.

\subsection{Implement Details}
We leverage the VGG16 and Resnet50 pre-trained in base categories as the feature extractors. Specifically,  the VGG16 and Resnet50 are leveraged as the  backbone of PSPNet~\cite{zhao2017pyramid}, and the abundant base-category samples are utilized to train the PSPNet. After this training process, the pre-trained backbones could  provide a generalized segmentation feature space for the following few-shot segmentation phase. The overall network is built on Pytorch.  The SGD optimizer with an initial learning rate 5e-3 is used for updating the parameters, and the training batch size is set to 4. To facilitate  the generalization ability of FRINet, the weights of the backbones are frozen, and only the weights of the rest network are learnable. Both Query and support images share the same-weight feature extractor and are propagated into the segmentation model with a size of $3 \times 256 \times 256$.  All the training images are
augmented with random horizontal flipping. For the 1-shot setting, the training phase lasts for 100 epochs, and the training process lasts for 50 epochs for the 5-shot setting. 
To better demonstrate the superiority of our proposed segmentation model,  a series of performance comparison experiments are performed in three splits of iSAID-$5^i$ at 1-shot and 5-shot settings.

\begin{table*}[t!]
\caption{Performance comparisons of different methods across different splits on the iSAID-$5^i$ dataset}
	\centering
	\scriptsize
	\footnotesize
	\renewcommand{\arraystretch}{1.3}
	\renewcommand{\tabcolsep}{6.0mm}		
	\scalebox{0.98}{
\begin{tabular}{clclclclclclclclcl}
\hline
\multicolumn{2}{c|}{\multirow{2}{*}{Method}} & \multicolumn{8}{c|}{1-shot}                                                                                      & \multicolumn{8}{c}{5-shot}                                                                                       \\ \cline{3-18} 
\multicolumn{2}{c|}{}                        & \multicolumn{2}{c}{Split-0} & \multicolumn{2}{c}{Split-1} & \multicolumn{2}{c}{Split-2} & \multicolumn{2}{c|}{mean} & \multicolumn{2}{c}{Split-0} & \multicolumn{2}{c}{Split-1} & \multicolumn{2}{c}{Split-2} & \multicolumn{2}{c}{mean}  \\ \hline
\multicolumn{18}{c}{VGG16}   \\ \hline
\multicolumn{2}{c|}{PANet~\cite{wang2019panet}}                   & \multicolumn{2}{c}{17.43}  & \multicolumn{2}{c}{11.43}  & \multicolumn{2}{c}{15.95}  & \multicolumn{2}{c|}{14.94} & \multicolumn{2}{c}{17.70}  & \multicolumn{2}{c}{14.58}  & \multicolumn{2}{c}{20.70}  & \multicolumn{2}{c}{17.66} \\
\multicolumn{2}{c|}{CANet~\cite{zhang2019canet}}                   & \multicolumn{2}{c}{19.73}  & \multicolumn{2}{c}{17.98}  & \multicolumn{2}{c}{30.93}  & \multicolumn{2}{c|}{22.88} & \multicolumn{2}{c}{23.45}  & \multicolumn{2}{c}{20.53}  & \multicolumn{2}{c}{30.12}  & \multicolumn{2}{c}{24.70} \\
\multicolumn{2}{c|}{PMMs~\cite{yang2020prototype}}                    & \multicolumn{2}{c}{20.87}  & \multicolumn{2}{c}{16.07}  & \multicolumn{2}{c}{24.65}  & \multicolumn{2}{c|}{20.53} & \multicolumn{2}{c}{23.31}  & \multicolumn{2}{c}{16.61}  & \multicolumn{2}{c}{27.43}  & \multicolumn{2}{c}{22.45} \\
\multicolumn{2}{c|}{PFENet~\cite{tian2020prior}}                  & \multicolumn{2}{c}{16.68}  & \multicolumn{2}{c}{15.30}  & \multicolumn{2}{c}{27.87}  & \multicolumn{2}{c|}{19.95} & \multicolumn{2}{c}{18.46}  & \multicolumn{2}{c}{18.39}  & \multicolumn{2}{c}{28.81}  & \multicolumn{2}{c}{21.89} \\
\multicolumn{2}{c|}{SDM~\cite{yao2021scale}}                     & \multicolumn{2}{c}{29.24}  & \multicolumn{2}{c}{20.80}  & \multicolumn{2}{c}{34.73}  & \multicolumn{2}{c|}{28.26} & \multicolumn{2}{c}{36.33}  & \multicolumn{2}{c}{27.98}  & \multicolumn{2}{c}{42.39}  & \multicolumn{2}{c}{35.57} \\
\multicolumn{2}{c|}{Ours}                    & \multicolumn{2}{c}{\textbf{42.86}}       & \multicolumn{2}{c}{\textbf{36.73}}       & \multicolumn{2}{c}{\textbf{42.51}}       & \multicolumn{2}{c|}{\textbf{40.70}}      & \multicolumn{2}{c}{\textbf{43.80}}       & \multicolumn{2}{c}{\textbf{37.18}}       & \multicolumn{2}{c}{\textbf{45.41}}       & \multicolumn{2}{c}{\textbf{42.13}}      \\ \hline
\multicolumn{18}{c}{Resnet50}    \\ \hline
\multicolumn{2}{c|}{PANet~\cite{wang2019panet}}                   & \multicolumn{2}{c}{12.36}  & \multicolumn{2}{c}{9.11}   & \multicolumn{2}{c}{12.05}  & \multicolumn{2}{c|}{11.17} & \multicolumn{2}{c}{13.82}  & \multicolumn{2}{c}{12.40}  & \multicolumn{2}{c}{19.12}  & \multicolumn{2}{c}{15.11} \\
\multicolumn{2}{c|}{CANet~\cite{zhang2019canet}}                   & \multicolumn{2}{c}{18.80}  & \multicolumn{2}{c}{15.62}  & \multicolumn{2}{c}{25.79}  & \multicolumn{2}{c|}{20.07} & \multicolumn{2}{c}{23.86}  & \multicolumn{2}{c}{18.54}  & \multicolumn{2}{c}{32.00}  & \multicolumn{2}{c}{24.80} \\
\multicolumn{2}{c|}{PMMs~\cite{yang2020prototype}}                    & \multicolumn{2}{c}{19.02}  & \multicolumn{2}{c}{18.51}  & \multicolumn{2}{c}{28.42}  & \multicolumn{2}{c|}{21.98} & \multicolumn{2}{c}{20.89}  & \multicolumn{2}{c}{20.87}  & \multicolumn{2}{c}{31.23}  & \multicolumn{2}{c}{24.33} \\
\multicolumn{2}{c|}{PFENet~\cite{tian2020prior}}                  & \multicolumn{2}{c}{18.75}  & \multicolumn{2}{c}{17.24}  & \multicolumn{2}{c}{22.09}  & \multicolumn{2}{c|}{19.36} & \multicolumn{2}{c}{19.57}  & \multicolumn{2}{c}{18.43}  & \multicolumn{2}{c}{26.14}  & \multicolumn{2}{c}{21.38} \\
\multicolumn{2}{c|}{SDM~\cite{yao2021scale}}                     & \multicolumn{2}{c}{34.29}  & \multicolumn{2}{c}{22.25}  & \multicolumn{2}{c}{35.62}  & \multicolumn{2}{c|}{30.72} & \multicolumn{2}{c}{39.88}  & \multicolumn{2}{c}{30.59}  & \multicolumn{2}{c}{45.70}  & \multicolumn{2}{c}{38.72} \\
\multicolumn{2}{c|}{Ours}                    & \multicolumn{2}{c}{\textbf{46.50}}       & \multicolumn{2}{c}{\textbf{36.94}}       & \multicolumn{2}{c}{\textbf{43.93}}       & \multicolumn{2}{c|}{\textbf{42.59}}      & \multicolumn{2}{c}{\textbf{48.85}}       & \multicolumn{2}{c}{\textbf{38.05}}       & \multicolumn{2}{c}{\textbf{46.46}}       & \multicolumn{2}{c}{\textbf{44.45}}      \\ \hline
\end{tabular}}
	\label{table1}	
\end{table*}

\begin{table*}[t!]
\caption{ Performance comparisons of diverse classes on the iSAID-$5^i$ dataset with 1-shot setting}
	\centering
        \normalsize
	\scalebox{0.9}{
\begin{tabular}{clccccccccccccccc}
\hline
\multicolumn{2}{c|}{Methods} & \multicolumn{1}{c|}{C1} & \multicolumn{1}{c|}{C2} & \multicolumn{1}{c|}{C3} & \multicolumn{1}{c|}{C4} & \multicolumn{1}{c|}{C5} & \multicolumn{1}{c|}{C6} & \multicolumn{1}{c|}{C7} & \multicolumn{1}{c|}{C8} & \multicolumn{1}{c|}{C9} & \multicolumn{1}{c|}{C10} & \multicolumn{1}{c|}{C11} & \multicolumn{1}{c|}{C12} & \multicolumn{1}{c|}{C13} & \multicolumn{1}{c|}{C14} & C15   \\ \hline
\multicolumn{17}{c}{VGG16}                                                                     \\ \hline
\multicolumn{2}{c|}{PANet~\cite{wang2019panet}}   & 13.85                   & 17.78                   & 19.78                   & 17.80                   & 17.94                   & 22.41                   & 11.75                   & 11.75                   & 6.36                    & 4.88                     & 12.96                    & 19.11                    & 23.49                    & 11.87                    & 12.33 \\
\multicolumn{2}{c|}{CANet~\cite{zhang2019canet}}   & 7.51                    & 12.87                   & 31.22                   & 30.45                   & 16.61                   & 18.31                   & 31.94                   & 23.15                   & 3.77                    & 12.55                    & 33.99                    & 51.41                    & 36.69                    & 17.61                    & 14.91 \\
\multicolumn{2}{c|}{PMMs~\cite{yang2020prototype}}    & 12.84                   & 23.06                   & 27.81                   & 26.65                   & 13.97                   & 18.48                   & 13.76                   & 25.58                   & 11.68                   & 10.88                    & 28.52                    & 34.62                    & 34.15                    & 16.32                    & 9.64  \\
\multicolumn{2}{c|}{PFENet~\cite{tian2020prior}}  & 3.94                    & 4.70                    & 30.95                   & 29.21                   & 14.59                   & 15.01                   & 14.63                   & 27.68                   & 10.13                   & 9.06                     & 29.40                    & 47.89                    & 25.23                    & 25.20                    & 11.63 \\
\multicolumn{2}{c|}{SDM~\cite{yao2021scale}}     & 26.55                   & \textbf{30.57}                   & 33.01                   & 35.06                   & 21.03                   & 20.54                   & 28.89                   & 32.45                   & 9.62                    & 12.59                    & \textbf{30.79}                    & 43.52                    & 49.47                    & 27.51                    & \textbf{22.37} \\
\multicolumn{2}{c|}{Ours} &\textbf{37.26}& 18.21 &\textbf{43.32} &\textbf{60.56} &\textbf{54.96} &\textbf{26.86} &\textbf{39.84} &\textbf{52.26} &\textbf{31.29} &\textbf{33.39} &28.57 &\textbf{65.20} &\textbf{51.08} 
 &\textbf{48.27} &19.41\\ \hline
\multicolumn{17}{c}{Resnet50}    \\ \hline
\multicolumn{2}{c|}{PANet~\cite{wang2019panet}}   & 7.27                    & 9.99                    & 14.49                   & 13.20                   & 16.85                   & 17.40                   & 10.23                   & 7.37                    & 6.21                    & 4.33                     & 8.35                     & 12.38                    & 21.28                    & 8.52                     & 9.75  \\
\multicolumn{2}{c|}{CANet~\cite{zhang2019canet}}   & 7.25                    & 27.25                   & 43.32                   & 6.17                    & 9.99                    & 6.16                    & 17.81                   & 28.58                   & 8.84                    & 16.73                    & 17.73                    & 56.97                    & 13.82                    & 27.41                    & 13.06 \\
\multicolumn{2}{c|}{PMMs~\cite{yang2020prototype}}    & 12.93                   & 17.92                   & 41.97                   & 7.43                    & 14.87                   & 13.24                   & 32.67                   & 26.08                   & 4.17                    & 16.37                    & 12.26                    & 61.49                    & 38.37                    & 17.92                    & 12.02 \\
\multicolumn{2}{c|}{PFENet~\cite{tian2020prior}}  & 2.98                    & 6.30                    & 33.29                   & 34.49                   & 16.71                   & 8.68                    & 8.67                    & 34.62                   & 18.19                   & 16.08                    & 28.58                    & 24.47                    & 7.37                     & 39.76                    & 10.31 \\
\multicolumn{2}{c|}{SDM~\cite{yao2021scale}}     & \textbf{37.66}                   & 34.37                   & 34.45                   & 39.81                   & 25.14                   & 16.77                   & 34.53                   & 30.50                   & 12.42                   & 17.02                    & 20.69                    & \textbf{56.83}                    & 42.80                    & 40.52                    & 17.26 \\
\multicolumn{2}{c|}{Ours} & 35.80  &\textbf{52.96}  &\textbf{43.75} & \textbf{53.36} & \textbf{46.61}
&\textbf{24.99}  &\textbf{48.87} & \textbf{63.38}  &\textbf{27.58}  &\textbf{19.89} &\textbf{34.30} &53.44 &\textbf{55.52} &\textbf{52.44} &\textbf{23.94}    \\ \hline
\end{tabular}}
\label{table2}	
\end{table*}

\subsection{Performance Analysis}
The comparisons between our proposed FRINet and other advanced FSS algorithms with the available source codes in the iSAID-$5^i$, dataset are shown in Table~\ref{table1}, and the best performance in each comparison is bolded. Clearly, our proposed FRINet achieves the best segmentation performance and could acquire better performance  with the reinforcement of backbones.  With VGG16 backbone, for the 1-shot setting, FRINet achieves 40.70 mean mIOU performance, which brings 12.44 $\%$ mIOU improvement. The best performance is achieved in the split0 with 42.86 mIOU performance. For the 5-shot setting, FRINet achieves 42.13 mean mIOU performance, which brings a  6.56 $\%$  mIOU improvement.  The best performance is achieved in the split2 with a 45.41 mIOU performance. Particularly, with both 1-shot and 5-shot settings, all splits could achieve more  than 3 $\%$  mIOU improvement. This suggests the FRINet could truly bring a splendid performance boost. Simultaneously, with the Resnet50 backbone, FRINet could achieve 42.59 mean mIOU performance for 1-shot setting, which brings 11.87 $\%$ mIOU improvement. Similar to the performance with VGG16, the best performance is obtained in the split0 with 46.50 mIOU performance. For the 5-shot setting, FRINet acquires 44.45 mean  mIOU performance, the best performance is also obtained in the spilt0 with 48.85 mIOU.  We could see that through our proposed FRINet, all splits could acquire an impressive performance boost. This phenomenon powerfully illustrates the effectiveness of the proposed FRINet and proves the FRINet successfully  encourages the improvement of few-shot aerial segmentation.

\begin{table*}[t!]
\caption{ Performance comparisons of diverse classes on the iSAID-$5^i$ dataset with 5-shot setting}
	\centering
        \normalsize
	\scalebox{0.9}{
\begin{tabular}{cllllllllllllllll}
\hline
\multicolumn{2}{c|}{Methods} & \multicolumn{1}{c|}{C1} & \multicolumn{1}{c|}{C2} & \multicolumn{1}{c|}{C3} & \multicolumn{1}{c|}{C4} & \multicolumn{1}{c|}{C5} & \multicolumn{1}{c|}{C6} & \multicolumn{1}{c|}{C7} & \multicolumn{1}{c|}{C8} & \multicolumn{1}{c|}{C9} & \multicolumn{1}{c|}{C10} & \multicolumn{1}{c|}{C11} & \multicolumn{1}{c|}{C12} & \multicolumn{1}{c|}{C13} & \multicolumn{1}{c|}{C14} & \multicolumn{1}{c}{C15} \\ \hline
\multicolumn{17}{c}{VGG16}    \\ \hline
\multicolumn{2}{c|}{PANet}   & 13.40                   & 17.79                   & 19.04                   & 18.11                   & 20.15                   & 22.97                   & 16.18                   & 15.65                   & 9.37                    & 8.73                     & 31.65                    & 21.63                    & 25.36                    & 12.70                    & 12.15                   \\
\multicolumn{2}{c|}{CANet}   & 11.87                   & 18.36                   & 33.26                   & 26.05                   & 27.69                   & 19.20                   & 29.93                   & 30.37                   & 11.76                   & 11.37                    & 23.76                    & 60.25                    & 28.67                    & 21.77                    & 16.16                   \\
\multicolumn{2}{c|}{PMMs}    & 14.99                   & 25.09                   & 28.49                   & 30.40                   & 17.58                   & 18.37                   & 13.40                   & 27.12                   & 12.45                   & 11.68                    & 35.46                    & 36.57                    & 37.08                    & 18.41                    & 9.63                    \\
\multicolumn{2}{c|}{PFENet}  & 2.64                    & 8.57                    & 29.09                   & 30.76                   & 21.26                   & 15.22                   & 21.58                   & 31.16                   & 12.06                   & 11.90                    & 33.56                    & 48.47                    & 24.92                    & 25.29                    & 11.79                   \\
\multicolumn{2}{c|}{SDM}     & 30.81                   & \textbf{38.59}                   & 37.34                   & 42.58                   & 32.35                   & \textbf{32.32}                   & 39.71                   & 37.23                   & 17.73                   & 12.93                    & 39.20                    & \textbf{66.78}                    & 39.06                    & 46.17                    & 20.72                   \\
\multicolumn{2}{c|}{Ours}    & \textbf{35.57} &8.01 &\textbf{58.46} &\textbf{68.03} &\textbf{48.96} &24.73 &\textbf{47.47} &\textbf{54.54} &\textbf{26.30} &\textbf{32.87} &\textbf{42.97} &56.07 &\textbf{53.19} &\textbf{51.55} &\textbf{23.25} \\ \hline
\multicolumn{17}{c}{Resnet50}   \\ \hline
\multicolumn{2}{c|}{PANet}   & 8.45                    & 11.68                   & 17.81                   & 14.22                   & 16.93                   & 18.86                   & 15.70                   & 9.40                    & 7.90                    & 10.13                    & 21.41                    & 24.94                    & 28.41                    & 9.80                     & 11.04                   \\
\multicolumn{2}{c|}{CANet}   & 32.47                   & 6.49                    & 39.8                    & 25.97                   & 14.59                   & 25.01                   & 25.27                   & 32.06                   & 1.31                    & 9.08                     & 26.99                    & 73.84                    & 33.83                    & 20.69                    & 4.69                    \\
\multicolumn{2}{c|}{PMMs}    & 13.44                   & 22.22                   & 42.12                   & 8.58                    & 18.06                   & 13.17                   & 37.88                   & 30.69                   & 6.41                    & 16.21                    & 14.62                    & 65.02                    & 42.99                    & 20.86                    & 12.66                   \\
\multicolumn{2}{c|}{PFENet}  & 10.13                   & 9.48                    & 30.71                   & 31.23                   & 16.31                   & 11.53                   & 14.15                   & 36.07                   & 15.58                   & 14.82                    & 38.52                    & 20.90                    & 20.41                    & 38.66                    & 12.22                   \\
\multicolumn{2}{c|}{SDM}     & \textbf{38.76}                   & 49.06                   & 50.06                   & 39.25                   & 22.26                   & 30.68                   & 45.34                   & 41.49                   & 20.21                   & 15.21                    & 32.61                    & \textbf{66.64}                    & \textbf{57.41}                    & 49.12                    & \textbf{22.71}                   \\
\multicolumn{2}{c|}{Ours}    &28.28  &\textbf{49.50}  &\textbf{50.40}  &\textbf{65.06}  &\textbf{51.01} &\textbf{33.84}  &\textbf{47.26}  &\textbf{57.01}  &\textbf{22.40}  &\textbf{29.74}  &\textbf{48.87} &58.15 &47.44 &\textbf{56.67} &21.20 \\ \hline
\end{tabular}}
\label{table3}	
\end{table*}

\begin{figure*}[t!]
  \begin{center}
  \includegraphics[width=1.0\linewidth]{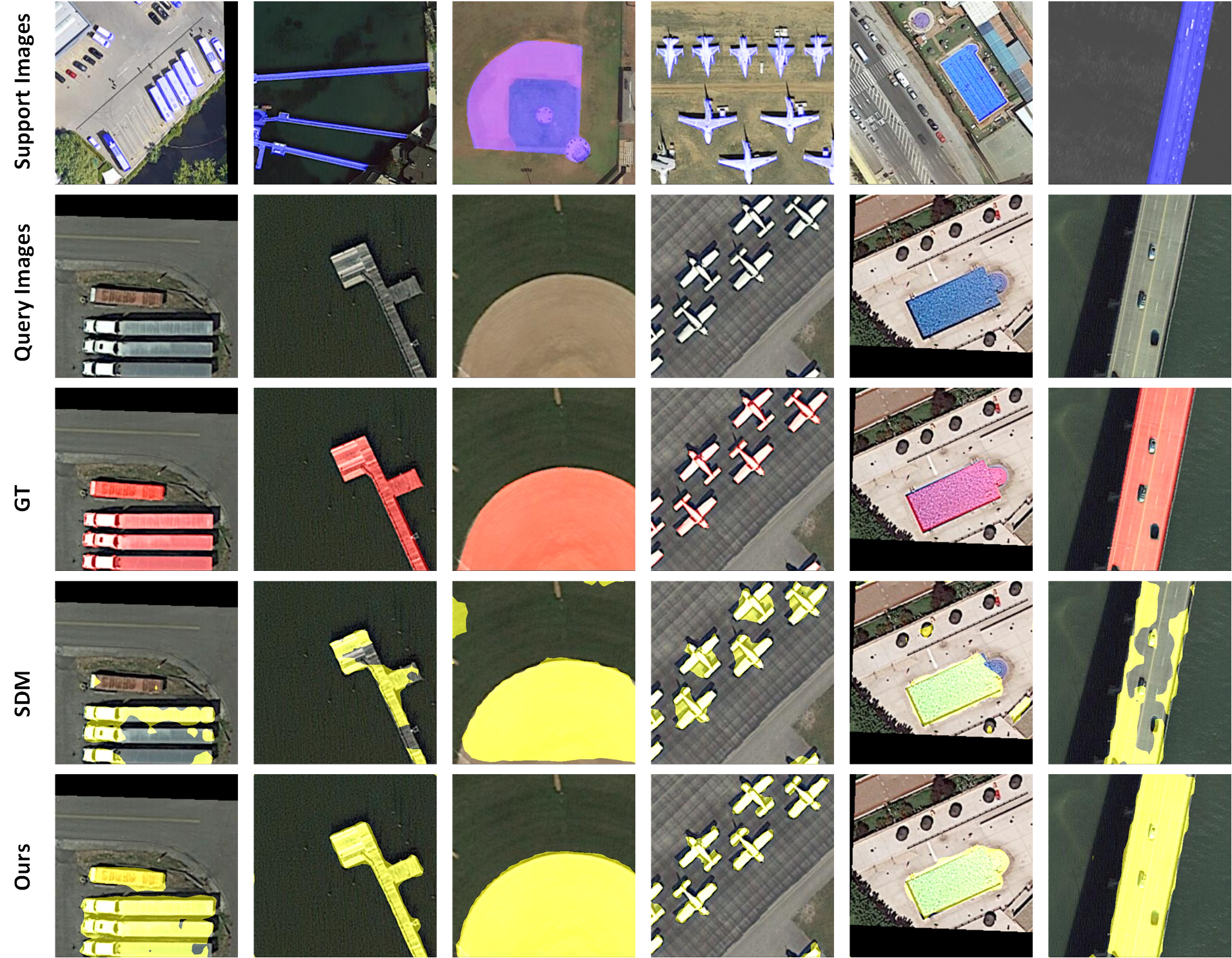}\\
  \caption{Qualitative results of proposed FRINet. From top to bottom: Support images, query images, the ground truth of query images, predictions of SDM network, and predictions of proposed FRINet. }\label{fig4}
  \end{center}
\end{figure*}

To further analyze the performance on diverse classes
with the few-shot setting, the detailed segmentation results are collected, and the results are respectively illustrated in Table~\ref{table2} and Table~\ref{table3}. As shown in Table~\ref{table2}, the proposed FRINet could achieve the best parsing performance in almost all categories. Particularly, the  best performance is 60.56 mIOU in the tennis court (C4) and 63.38 mIOU in the large vehicle (C8) respectively for the VGG16 and Resnet50 backbones. Moreover, the best improvement is 33.93 $\%$ mIOU improvement in the basketball court (C5) and 32.88 $\%$ mIOU  improvement in the large vehicle (C8), respectively, for the VGG16 and Resnet50 backbones. These results suggest that the rotation invariance is very critical for these categories, and the proposed FRINet successfully mines the rotation invariance.  However, our proposed FRINet still fails in some categories. For instance, a huge performance gap could be seen in the storage tank (C2). This phenomenon demonstrates that the proposed FRINet may still have limited ability to mine the rotation invariance of the small-scale aerial objects like the storage tank. 
\begin{table}[t]
	\centering
	\scriptsize
	\footnotesize
	\renewcommand{\arraystretch}{1.0}
	\renewcommand{\tabcolsep}{4.0mm}
	\caption{Ablation study of the newly provided backbones. NB denotes newly provided backbones. }
	\scalebox{1.0}{
\begin{tabular}{cl|cccc}
\hline
\multicolumn{2}{c|}{Methods}     & Split-0        & Split-1        & Split-2        & Mean           \\ \hline
\multicolumn{2}{c|}{Baseline}     & 29.24          & 20.80          & 34.73          & 28.26          \\
\multicolumn{2}{c|}{Baseline+NB} & \textbf{40.33} & \textbf{34.65} & \textbf{39.07} & \textbf{38.01} \\ \hline
\end{tabular}}
	\label{table4}
\end{table}

\begin{table}[t]
	\centering
	\scriptsize
	\footnotesize
	\renewcommand{\arraystretch}{1.0}
	\renewcommand{\tabcolsep}{4.0mm}
	\caption{Ablation study of the rotation-adaptive matching and rotation-invariant segmentation. RAM illustrates the rotation-adaptive matching, RIS denotes the rotation-invariant segmentation. }
	\scalebox{1.0}{
		\begin{tabular}{cc|cccc}
			\hline
			\multicolumn{1}{c}{\multirow{1}{*}{RAM}}&  \multicolumn{1}{c|}{RIS} &\multicolumn{1}{c}{Split-0}  &\multicolumn{1}{c}{Split-1}  &\multicolumn{1}{c}{Split-2} &\multicolumn{1}{c}{Mean} \\
			\hline
			\multicolumn{1}{c}{} &    \multicolumn{1}{c|}{}  &40.33 &34.65 &39.07 &38.01  \\ 
			\multicolumn{1}{c}{\checkmark} & \multicolumn{1}{c|}{} & 41.44&35.04 & 41.91 & 39.46  \\
   			\multicolumn{1}{c}{} & \multicolumn{1}{c|}{\checkmark} &40.75 &35.07 &41.72 &39.18  \\
			\multicolumn{1}{c}{\checkmark} &  \multicolumn{1}{c|}{\checkmark} &\textbf{42.86} &\textbf{36.73} &\textbf{42.51} &\textbf{40.70}  \\  \hline		
	\end{tabular}}
	\label{table5}
\end{table}
Focusing on the performance in Table~\ref{table3}, the proposed FRINet still  obtains the best segmentation performance in almost all categories. Specifically, the best performance are 68.03 mIOU and 65.06 mIOU in the tennis court (C4) respectively for the VGG16 and Resnet50 backbones. Furthermore, the best performance boost is also achieved in the tennis court (C4), i.e., 25.45 $\%$ and 25.81 $\%$ mIOU improvement. This phenomenon again tells us the rotation invariance of the tennis court (C4) is well-mined by the FRINet. However, similarly, the performance in small-scale aerial objects like the storage tank is still very limited. We believe the main reason is that the resolution of the features of small-scale aerial objects is also very small, thus the change of rotation could not call a string-enough feature change to support the FRINet to mine the inner rotation invariance. This phenomenon could inspire further work to focus on designing an appropriate method to mine the rotation-invariance information-limited small-scall aerial objects.

To further directly analyze the performance of the proposed FRINet, the qualitative results are shown in Figure~\ref{fig4}. Obviously, the pleasant segmentation performance successfully demonstrates the effectiveness of the proposed FRINet. Particularly, rotation-adaptive matching could adaptively activate aerial objects with varying orientations, and the rotation-invariant segmentation helps the network eliminate the background and  recognize more missing foreground. For instance, in the first line, the orientation of large vehicles in the support image is totally different from the orientation of large vehicles in the query image. Thus, the previous method fails to recognize these  orientation-changed objects. But our proposed FRINet successfully recognizes these orientation-varying objects. Furthermore, we could see a similar phenomenon in the second row (the harbor) and in the fifth row (the swimming pool). 

Interestingly, for the fourth row( the airplane category), we could find that our proposed FRINet could eliminate some very tiny but essential backgrounds. We believe these impressive results benefit from the rotation-invariant segmentation.  The rotation-invariant segmentation encourages the network to make consistent predictions from diverse orientations and further fuses these predictions in a complementary manner. Apparently, this  manner could help the network to accumulate the segmentation results in different views, which could produce a more precise parsing result from coarse results.

\subsection{Ablation Study}
To evaluate the effectiveness of our proposed FRINet, a series of experiments are performed to analyze the effect of the key components of the FRINet. All ablation experiments are conducted under the VGG16 backbone and 1-shot setting.

First, the effect of the newly provided backbones is studied. As shown in Table~\ref{table4}, the newly provided backbone brings very impressive performance improvement. For split0 and split1, it provides more than 10$\%$ mIOU boost. For the split2, it also brings nearly 5$\%$ mIOU improvement. Overall, it gives 9.76$\%$ performance improvement. All these improvements prove the newly provided backbones truly provide a better feature space for the few-shot aerial segmentation task.

Then, based on the newly provided backbones, we further study the effect of the proposed rotation-adaptive matching and the rotation-invariant segmentation, i.e., the rotation-varying  segmentation results supervised by the same ground truth are fused to obtain the final parsing result in a complementary manner. As shown in Table~\ref{table5}, RAM denotes the rotation-adaptive matching, and RIS denotes the rotation-invariant segmentation.  The rotation-adaptive matching could bring  a performance boost for splits. Particularly, rotation-adaptive matching leads to the best performance enhancement, namely 2.84$\%$ mIOU improvement for split2. This result implies that more orientation-varying aerial objects are not activated for split2 in previous FSS solutions. For the rotation-invariant segmentation, the best performance improvement, namely 2.64$\%$ mIOU improvement,  is also obtained in the split2. This phenomenon tells us that some orientation-varying aerial objects are falsely recognized. By jointly leveraging RAM and RIS, the proposed FRINet successfully achieves 40.70 mIOU performance. These performance improvements powerfully illustrate that the proposed rotation-adaptive matching and the rotation-invariant segmentation could efficiently help the network parse the rotation-varying aerial objects.

\begin{table}[t]
	\centering
	\scriptsize
	\footnotesize
	\renewcommand{\arraystretch}{1.0}
	\renewcommand{\tabcolsep}{4.0mm}
	\caption{The difference between the proposed FRINet and the traditional rotation data augmentation. w/o denotes without, w denotes with, and Aug means the rotation data augmentation. }
	\scalebox{1.0}{
\begin{tabular}{cl|cccc}
\hline
\multicolumn{2}{c|}{Methods} & Split-0        & Split-1        & Split-2        & Mean           \\ \hline
\multicolumn{2}{c|}{w/o Aug} & 39.33          & 33.36          & 38.33          & 37.01          \\
\multicolumn{2}{c|}{w Aug}   & 40.33          & 34.65          & 39.07          & 38.01          \\
\multicolumn{2}{c|}{Ours}    & \textbf{42.86} & \textbf{36.73} & \textbf{42.51} & \textbf{40.70} \\ \hline
\end{tabular}}
	\label{table6}
\end{table}

\begin{table}[t]
	\centering
	\scriptsize
	\footnotesize
	\renewcommand{\arraystretch}{1.0}
	\renewcommand{\tabcolsep}{4.0mm}
	\caption{The ablation study of orientations. }
	\scalebox{1.0}{
\begin{tabular}{cl|cccc}
\hline
\multicolumn{2}{c|}{Orientations}      & Split-0        & Split-1        & Split-2        & Mean           \\ \hline
\multicolumn{2}{c|}{$[0^{\circ}]$}            & 41.44          & 35.04          & 41.91          & 39.46          \\
\multicolumn{2}{c|}{$[0^{\circ}$,$90^{\circ}]$}         & 41.98          & 35.99          & 42.11          & 40.03          \\
\multicolumn{2}{c|}{[$0^{\circ}$,$90^{\circ}$,$180^{\circ}]$}     & 42.36          & 36.43          & 42.39          & 40.39          \\
\multicolumn{2}{c|}{[$0^{\circ}$,$90^{\circ}$,$180^{\circ}$,$270^{\circ}]$} & \textbf{42.86} & \textbf{36.73} & \textbf{42.51} & \textbf{40.70} \\ \hline
\end{tabular}}
	\label{table7}
\end{table}

To distinguish our proposed FRINet from the traditional rotation data augmentation, some ablation experiments are conducted, and the experimental results are shown in Table~\ref{table6}. The traditional rotation data augmentation randomly rotates the images at a limited angle. Though this strategy  could provide different orientation-views for training, only a single orientation-view is leveraged in the query-support matching. The orientation-varying objects could still not be activated and be falsely recognized. Moreover, as the experimental results imply, our FRINet could bring a more impressive performance boost. Thus, our FRINet is totally different from the traditional rotation data augmentation strategy.

In our FRINet, rotated images with three angles ($90^\circ$,$180^\circ$,$270^\circ$) and original images are adopted in our FRINet. To demonstrate the necessity of diverse angles, some ablation experiments are performed and results are shown in Table~\ref{table7}. Clearly, with the addition of orientations, all splits could an increasing of mIOU performance, and the best performance is obtained when all orientations are included in the network. This phenomenon illustrates that more orientations could help FRINet parse more orientation-varying objects.

\section{Conclusion}
In this paper, we propose a novel few-shot rotation-invariant aerial semantic segmentation network. Particularly, aiming at activating aerial objects with varying orientations,  orientation-varying yet category-consistent support information is leveraged to perform the rotation-adaptive matching. Meanwhile, rotation-varying segmentation results supervised by the same ground truth are fused to obtain the final rotation-invariant parsing result in a complementary manner. Additionally, the backbones pre-trained in the base categories are newly provided to offer a better feature space. The extensive experiments in the few-shot aerial semantic segmentation benchmark demonstrate  our model achieves state-of-the-art performances.




%





\ifCLASSOPTIONcaptionsoff
  \newpage
\fi





\bibliographystyle{IEEEtran}
\bibliography{IEEEabrv,Bibliography}

\vfill


\end{document}